\PassOptionsToPackage{dvipsnames}{xcolor}
\documentclass[review]{elsarticle}
\makeatletter
\def\ps@pprintTitle{%
 \let\@oddhead\@empty
 \let\@evenhead\@empty
 \def\@oddfoot{}%
 \let\@evenfoot\@oddfoot}
\makeatother
\usepackage{tikz}
\usepackage{comment}
\usepackage{amsmath,amssymb} 
\usepackage{color}
\usepackage{epsfig}
\usepackage[inline]{enumitem}
\usepackage[caption=false]{subfig}
\usepackage{soul}
\usepackage[utf8]{inputenc} 
\usepackage[T1]{fontenc}    
\usepackage{booktabs}       
\usepackage{amsfonts}       
\usepackage{nicefrac}       
\usepackage{microtype}      
\usepackage{algorithmic}
\usepackage{algorithm}
\usepackage{numprint}
\usepackage{siunitx}
\usepackage{etoolbox}
\newcommand{\ubold}{\fontseries{b}\selectfont}
\robustify\ubold
\usepackage{bbold}
\usepackage{wrapfig}
\usepackage{bm}
\usepackage[caption=false]{subfig}
\usepackage{float}
\usepackage{pifont}

\usepackage{lineno,hyperref}
\usepackage{array,multirow,graphicx}

\usepackage[dvipsnames]{xcolor}
\makeatletter
\newcommand\footnoteref[1]{\protected@xdef\@thefnmark{\ref{#1}}\@footnotemark}
\makeatother

\modulolinenumbers[5]

\journal{Journal of Pattern Recognition}

\begin{document}

\pagenumbering{gobble}

\let\WriteBookmarks\relax
\def\floatpagepagefraction{1}
\def\textpagefraction{.001}
\pagenumbering{arabic}

\newcommand{\ttinneriter}{\texttt{inner\_iters}}
\newcommand{\ttit}{\texttt{iter}}
\newcommand{\ttmaxiter}{\texttt{max\_iter}}
\newcommand{\ttbestScore}{\texttt{best\_score}}
\newcommand{\ttbestR}{\texttt{best\_rotation}}
\newcommand{\ttbestT}{\texttt{best\_translation}}

\newcommand{\cL}{\mathcal{L}}
\newcommand{\cM}{\mathcal{M}}
\newcommand{\cN}{\mathcal{N}}
\newcommand{\cI}{\mathcal{I}}
\newcommand{\cS}{\mathcal{S}}
\newcommand{\cD}{\mathcal{D}}
\newcommand{\cP}{\mathcal{P}}
\newcommand{\cQ}{\mathcal{Q}}
\newcommand{\cO}{\mathcal{O}}
\newcommand{\cT}{\mathcal{T}}
\newcommand{\cad}{\mathcal{d}}
\newcommand{\cX}{\mathcal{X}}
\newcommand{\cXh}{\hat{\mathcal{X}}}
\newcommand{\cC}{\mathcal{C}}
\newcommand{\cF}{\mathcal{F}}

\newcommand{\be}{\mathbf{e}}
\newcommand{\br}{\mathbf{r}}
\newcommand{\bx}{\mathbf{x}}
\newcommand{\bxh}{\hat{\mathbf{x}}}
\newcommand{\bX}{\mathbf{X}}
\newcommand{\bY}{\mathbf{Y}}
\newcommand{\bZero}{\mathbf{0}}
\newcommand{\hbX}{\hat{\mathbf{X}}}
\newcommand{\bS}{\mathbf{S}}
\newcommand{\bs}{\mathbf{s}}
\newcommand{\bp}{\mathbf{p}}
\newcommand{\bq}{\mathbf{q}}
\newcommand{\bD}{\mathbf{D}}
\newcommand{\bd}{\mathbf{d}}
\newcommand{\bA}{\mathbf{A}}
\newcommand{\bR}{\mathbf{R}}
\newcommand{\bt}{\mathbf{t}}
\newcommand{\bH}{\mathbf{H}}
\newcommand{\bh}{\mathbf{h}}
\newcommand{\by}{\mathbf{y}}
\newcommand{\bz}{\mathbf{z}}
\newcommand{\bu}{\mathbf{u}}
\newcommand{\ba}{\mathbf{a}}
\newcommand{\bg}{\mathbf{g}}
\newcommand{\bo}{\mathbf{o}}
\newcommand{\bl}{\mathbf{l}}
\newcommand{\bOnes}{\mathbf{1}}
\newcommand{\bF}{\mathbf{F}}
\newcommand{\bK}{\mathbf{K}}
\newcommand{\bI}{\mathbf{I}}
\newcommand{\tdf}{\tilde{f}}
\newcommand{\tdh}{\tilde{h}}

\newcommand{\bbR}{\mathbb{R}}
\newcommand{\bbE}{\mathbb{E}}
\newcommand{\bbD}{\mathbb{D}}
\newcommand{\bbF}{\mathbb{F}}
\newcommand{\bbFh}{\hat{\mathbb{F}}}
\newcommand{\bmu}{\boldsymbol{\mu}}
\newcommand{\bhr}{\hat{\mathbf{r}}}
\newcommand{\bJ}{\mathbf{J}}
\newcommand{\Nsample}{$N_\text{sample}$}

\newcommand{\kernel}{\psi}

\newcommand{\residual}{\mathbf{r}}
\newcommand{\btheta}{\boldsymbol{\theta}}

\newcommand{\cV}{\mathcal{V}}
\newcommand{\cE}{\mathcal{E}}
\newcommand{\cB}{\mathcal{B}}
\newcommand{\cG}{\mathcal{G}}
\newcommand{\etal}{\textit{et al.}}
\newcommand{\xmark}{\ding{55}}%
\newcommand{\cmark}{\ding{51}}%

\begin{frontmatter}

    \title{A Generically Contrastive Spatiotemporal Representation Enhancement for 3D Skeleton Action Recognition}

    \author[bupt]{Shaojie Zhang}
    \author[bupt]{Jianqin Yin\corref{mycorrespondingauthor}}
    \cortext[mycorrespondingauthor]{Corresponding author}
    \ead{jqyin@bupt.edu.cn}
    \author[bupt]{Yonghao Dang}
    \address[bupt]{School of Artificial Intelligence, Beijing University of Posts and Telecommunications, \\ Beijing, China}


    \begin{abstract}
        Skeleton-based action recognition is a central task in computer vision and human-robot interaction. However, most previous methods suffer from overlooking the explicit exploitation of the latent data distributions (\textit{i.e.}, the intra-class variations and inter-class relations), thereby leading to confusion about ambiguous samples and sub-optimum solutions of the skeleton encoders. To mitigate this, we propose a \textbf{C}ontrastive \textbf{S}patiotemporal \textbf{R}epresentation \textbf{E}nhancement (CSRE) framework to obtain more discriminative representations from the sequences, which can be incorporated into various previous skeleton encoders and can be removed when testing. Specifically, we decompose the representation into spatial- and temporal-specific features to explore fine-grained motion patterns along the corresponding dimensions. Furthermore, to explicitly exploit the latent data distributions, we employ the attentive features to contrastive learning, which models the cross-sequence semantic relations by pulling together the features from the positive pairs and pushing away the negative pairs. Extensive experiments show that CSRE with five various skeleton encoders (HCN, 2S-AGCN, CTR-GCN, Hyperformer, and BlockGCN) achieves solid improvements on five benchmarks. The code will be released at \url{https://github.com/zhshj0110/CSRE}.
    \end{abstract}
\end{frontmatter}

\section{Introduction}
Human action recognition is an active task in the computer vision area. Due to its wide range of applications in human-computer interaction, video analysis, virtual reality, and so on, this task has been researched extensively in the past decade \cite{ren2024survey}. In recent years, with the development of depth sensors \cite{cao2017realtime} and human pose estimation algorithms \cite{dang2019deep}, the skeleton sequence consisting of coordinates of human joints can be easily acquired. Skeleton-based action recognition utilizing human joints has attracted much interest due to its robustness to background clutter, illumination, and viewpoint changes \cite{ren2024survey}.

\begin{figure}
    \centering
    \includegraphics[scale=0.58]{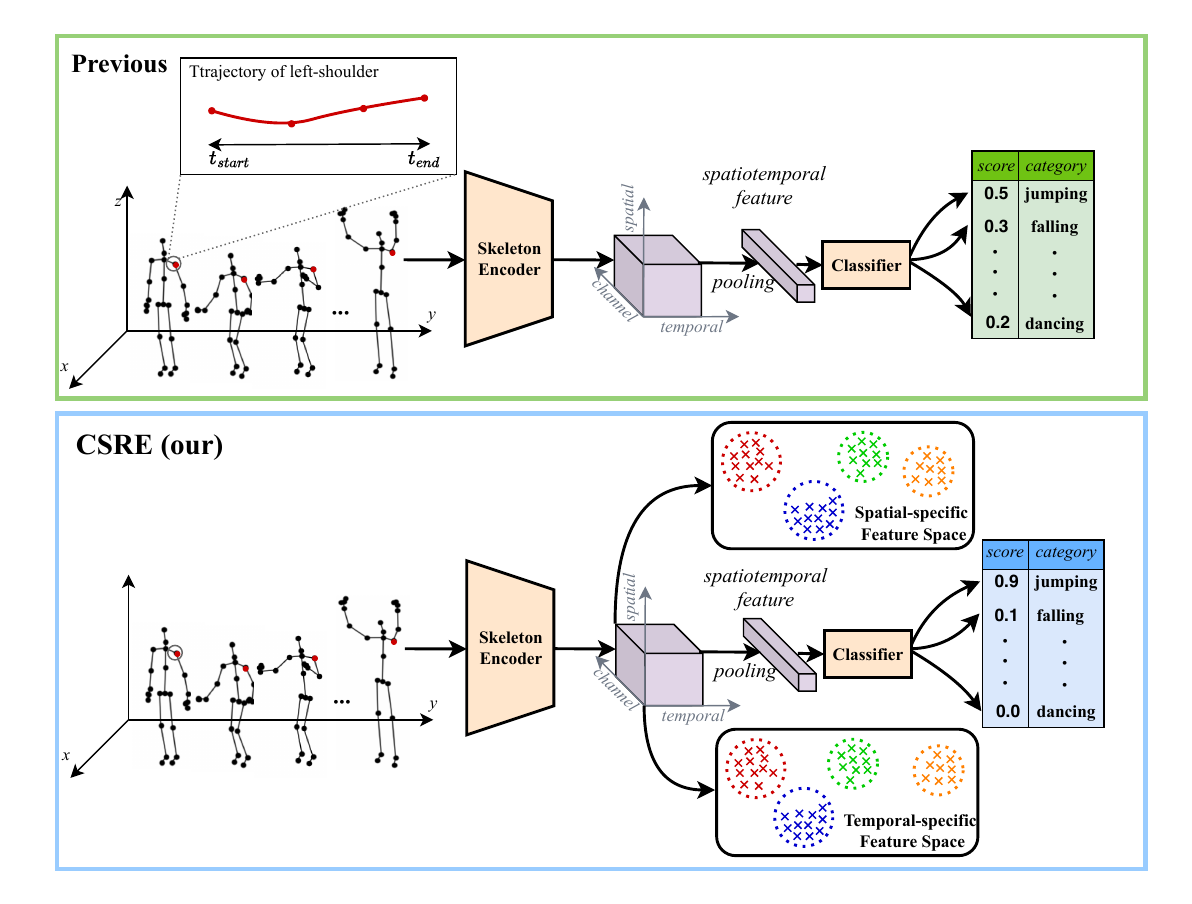}
    \caption{Comparison between previous method (in green) and proposed CSRE framework (in blue). Compared to the previous method, our CSRE can enable the skeleton encoder better to explore the fine-grained motion patterns along the spatial and temporal dimensions, which is beneficial for recognizing ambiguous action samples.}
    \label{fig: Motivation}
\end{figure}

Generally, most of the previous approaches \cite{shi2019two,chen2021channel,chen2021multi} in skeleton-based action recognition follow a framework as shown in Figure \ref{fig: Motivation}. A skeleton encoder aggregates the spatiotemporal co-occurrence feature \cite{li2018co}, and a softmax-based linear classifier is used to project the pooling feature to the category distributions. However, this widespread framework has some limitations. The main problem is the absence of explicit exploitation of the latent data distributions, thereby leading to confusion about ambiguous samples and sub-optimum solutions of skeleton encoders.

\begin{figure*}
    \centering
    \includegraphics[scale=0.145]{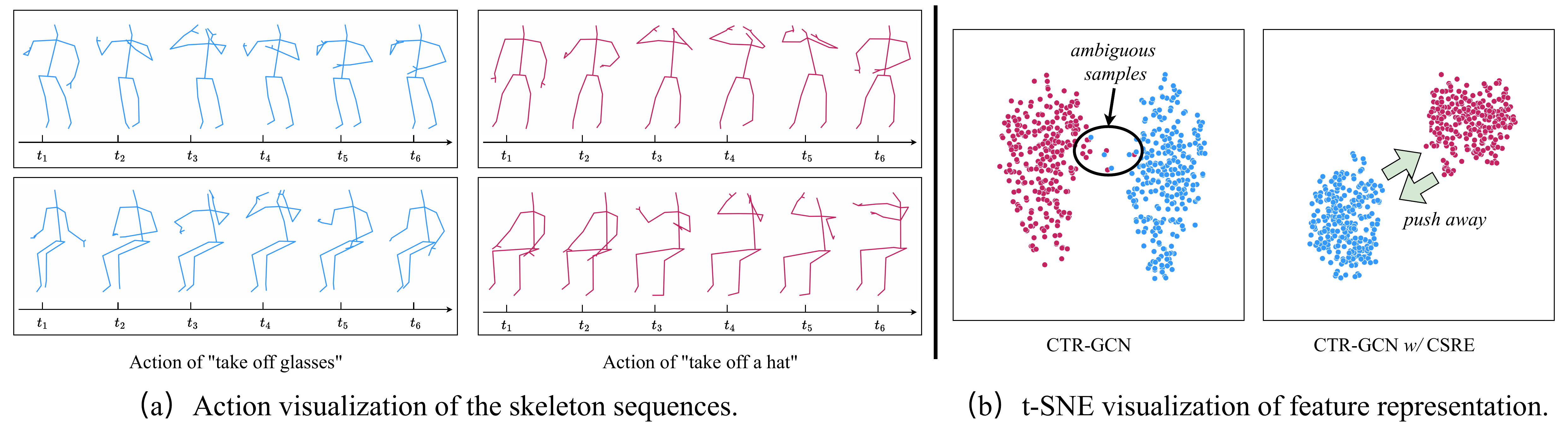}
    \caption{Visualization of samples from two confused classes (``take off glasses" (blue) and ``take off a hat" (red). (a) Visualization of sequences performed by different persons (each row represents a different person) in the same view. (b) Visualization of feature representation by t-SNE for the corresponding action in the test set. The left is from the CTR-GCN, while the right is from our method. Each color denotes a certain class.}
    \label{fig: vis}
\end{figure*}

Specifically, because the important interactive objects and contexts are omitted in skeletonization, it is difficult to distinguish the action categories with similar motion patterns significantly. For example, as shown in Figure \ref{fig: vis} (a), the action ``take off glasses" and the action ``take off a hat" exhibit a high degree of similarity due to the absence of the interactive objects (the hat and glasses). Furthermore, due to individual differences and the unknowability of the external environment, actions within the same category exhibit a wide range of intra-class variation. Although previous skeleton encoders extract discriminative representations from the skeleton sequences for classification, the linear classifier is typically trained to optimize accuracy only, paying less attention to exploiting the latent data distributions \cite{wang2023visual}. The t-SNE \cite{van2008visualizing} visualization of features extracted from CTR-GCN \cite{chen2021channel} is shown in Figure \ref{fig: vis} (b), which can't locate a distinguishable classification boundary because of the ambiguous samples. Therefore, we suppose explicit modeling of the intra-class variation and inter-class relation is crucial for recognizing the ambiguous samples.

To alleviate this drawback, As shown in \ref{fig: Motivation}, we propose a generically \textbf{C}ontrastive \textbf{S}patiotemporal \textbf{R}epresentation \textbf{E}nhancement (CSRE) framework to encourage the skeleton encoders to learn more discriminative representations to distinguish ambiguous samples. It's worth noting that our CSRE can be combined with previous skeleton encoders and removed naturally without additional consumption when testing. In the \textit{training} phase, to enable the network to better explore the fine-grained motion patterns among ambiguous actions along the topological and temporal dimensions, the spatiotemporal representation is decomposed into spatial and temporal components using the designed spatial-temporal feature decomposition (STFD) module. Moreover, we build two memory banks to store the spatial-specific and temporal-specific features, respectively, and sample the positives and negatives from them according to labels. In particular, we employ contrastive learning to model the cross-sequence semantic relations by pulling together the decomposed features from positive pairs and pushing away the features from negative pairs. With constant contrast, the skeleton encoders are forced to explicitly explore the latent data distributions with distinct inter-class relations and reduced intra-class variations. Consequently, in the \textit{testing} phase, the skeleton encoders can directly predict the categories with higher accuracy.

We summarize our main contributions as follows:
\begin{itemize}
    \item To extract more discriminative representation to distinguish the ambiguous samples, we proposed a novel and generically CSRE framework to decompose the features into spatial- and temporal-specific features and apply them for contrastive learning to explore the latent data distributions explicitly.

    \item The proposed CSRE can be seamlessly incorporated into various previous skeleton encoders, which can be regarded as plug-and-play in the training stage and can be removed at the testing stage.

    \item Extensive experiments show that CSRE achieves significant improvements based on five various state-of-the-art methods (HCN \cite{li2018co}, 2S-AGCN \cite{shi2019two}, CTR-GCN \cite{chen2021channel}, Hyperformer \cite{zhou2022hypergraph}, and BlockGCN \cite{zhou2024blockgcn}) on five benchmarks.
\end{itemize}

\section{Related Work}
\subsection{Skeleton-based Action Recognition}
Skeleton-based action recognition aims to classify actions from sequences of human keypoints. In early studies, Recurrent neural network (RNN) based methods \cite{du2015hierarchical, song2017end} and Convolution neural network (CNN) based methods \cite{li2018co, xu2022topology, dang2020dwnet} are popular choices for solving the skeleton-based action recognition problem.  However, both CNN-based and RNN-based methods overlook human topology and the spatial interactions between joints.

The human body can be abstracted as a graph structure, which treats the joint as a node and the bone between two joints as an edge. Thus, GCN-based methods were introduced to this task. ST-GCN \cite{yan2018spatial} adopted GCN on the predefined spatial-temporal graphs to model the relations between joints. 2S-AGCN \cite{shi2019two} modeled the correlation between two joints given corresponding features with the self-attention mechanism. CTR-GCN \cite{chen2021channel} proposed a channel-wise topology refinement graph convolution to model fine-grained relations. InfoGCN \cite{chi2022infogcn} combined a learning objective and an encoding method to break the information bottleneck. HD-GCN \cite{lee2023hierarchically} effectively constructs an HD-Graph by decomposing every joint node into several sets to extract major adjacent and distant edges. DD-GCN \cite{li2023dd} constructs the directed diffusion graph for action modeling and introduces the activity partition strategy. BlockGCN \cite{zhou2024blockgcn} introduces BlockGC to streamline multi-relational modeling, reducing parameters while improving performance over traditional GCNs.

With the popularity of transformers in computer vision, transformer-based methods have also been investigated for this task. ST-TR \cite{plizzari2021skeleton} proposed a two-stream transformer architecture to model the spatial and temporal dimensions, respectively. Hyperformer \cite{zhou2022hypergraph} applied transformer to model spatial-temporal features from the hyper-graph of the sequences. ISTA-Net \cite{wen2023interactive} introduces the transformer to model spatial, temporal, and interactive relations simultaneously. STC-Net \cite{lee2023leveraging} proposes the novel STC module and Dilated Kernels for Graph Convolution, enabling adaptive spatiotemporal coverage and long-range dependency modeling without increasing parameters. SelfGCN \cite{wu2024selfgcn} combines graph convolution and self-attention via the MFSG module for modeling both local and global joint relationships and the TSSA module for learning temporal-specific spatial features, addressing limitations in long-range dependency modeling and uniform skeleton topology.

\subsection{Spatiotemporal Feature Decomposition}
Human action has the property of spatial-temporal coupling. Thus, previous methods \cite{li2020spatio,plizzari2021skeleton,gedamu2023relation} tried to decompose features into specific dimensions and utilized the dual-branch or sequential network architecture to extract temporal-spatial features. STA \cite{li2020spatio} proposed a spatiotemporal attention module that simultaneously learns frame-level and channel-level attention weights to enhance 3D CNNs' ability to capture discriminative spatial and temporal features for video-based action recognition. ST-TR \cite{plizzari2021skeleton} proposed a novel two-stream Transformer-based model that employs Spatial Self-Attention (SSA) and Temporal Self-Attention (TSA) modules to dynamically model spatial and temporal dependencies in skeleton data, thereby effectively decomposing and capturing the spatial and temporal features for action recognition. RSA-Net \cite{gedamu2023relation} proposed a Relation-mining Self-Attention Network that decomposes spatiotemporal features by using pairwise self-attention to capture joint relationships and unary self-attention to model global dependencies while also separately computing action semantics and positional information with distinct projection matrices. Although these previous methods have attempted to decompose features into spatial and temporal-specific features, they still lack the explicit modeling of the latent data distributions along the specific dimensions.

\subsection{Contrastive Learning in Skeleton-based Action Recognition}
Recently, contrastive learning has been widely used in self-supervised representation learning. The core idea is to learn the discriminative features by pulling the positive pairs and pushing away the negative pairs in the embedded space. The positive pairs are features from transformed sample versions with different augmentations, and the negative pairs are from different samples. InstDisc \cite{wu2018unsupervised} maintained a memory bank for storing representations. SimCLR \cite{chen2020simple} proposed adopting a huge batch size to learn more useful representations. MoCo \cite{he2020momentum} built a dynamic dictionary with a momentum update based on a memory bank to keep the stored representations consistent. Inspired by this, our work tries to introduce memory bank mechanisms for contrastive learning to obtain discriminative features.

Contrastive learning was usually adopted in prior works \cite{xu2023spatiotemporal, guo2024improving, dong2023hierarchical} to learn the invariance from the skeleton sequences for unsupervised learning in this field. Moreover, several works are applying contrastive learning to supervised action recognition. GAP \cite{xiang2023generative} proposed a framework for contrasting the features from visual and text modalities to enhance the representation by using knowledge about actions and human body parts. SkeletonGCL \cite{huang2023graph} explicitly explored the rich cross-sequence relations by using graph contrastive learning. However, SkeletonGCL can only be incorporated into the GCN-based methods, which limits its generalization. FR-Head \cite{zhou2023learning} employed multi-level contrastive learning in GCN-based methods to distinguish ambiguous actions. However, the methods mentioned above are more like surgery for GCN-based skeleton encoders, leading to limitations on generalizability to other types of methods.

\section{Methodology}
In this section, we elaborate on the structure and training methodology of CSRE. Firstly, we state the problem formulation of skeleton-based action recognition and the generic modeling progress of previous approaches in Section \ref{Preliminaries}. Subsequently, we introduce the structure of the proposed STFD module in Section \ref{STFD}. Finally, we present the overall framework and training objective in Section \ref{CSRE}.

\subsection{Preliminaries}
\label{Preliminaries}
Skeleton-based action recognition is to predict the categories from the input sequences. We define the human skeleton sequence with $J$ keypoints and $T_{0}$ frames in 3D space as $\boldsymbol{S} \in \mathbb{R}^{J \times T_{0} \times 3}$. The sequence is fed into a skeleton encoder $\mathcal{F}$ to extract the features $\boldsymbol{X} \in \mathbb{R}^{J \times T \times C}$. The skeleton encoder aggregates spatial and temporal information to obtain discriminative spatial-temporal features. After fully spatial-temporal aggregations, a global pooling (GP) layer is adapted to summarize the global features. Finally, a softmax-based fully-connected (FC) layer maps the global feature to a probability prediction $\hat{\boldsymbol{y}} \in \mathbb{R}^{K}$ of $K$ candidate categories. The process can be defined as:

\begin{equation}
    \boldsymbol{X} = \mathcal{F}(\boldsymbol{S})
\end{equation}
\begin{equation}
    \boldsymbol{y} = FC(GP(\boldsymbol{X}))
\end{equation}
In the process of optimization, a standard cross-entropy loss is applied to supervise the prediction distribution and the ground truth $\boldsymbol{y}$ as follows:
\begin{equation}
    \mathcal{L}_{CE} = -\sum_{i} \boldsymbol{y}_i log \hat{\boldsymbol{y}_i}
\end{equation}

\subsection{Spatial Temporal Feature Decomposition}
\label{STFD}
\begin{figure*}
    \centering
    \includegraphics[scale=0.6]{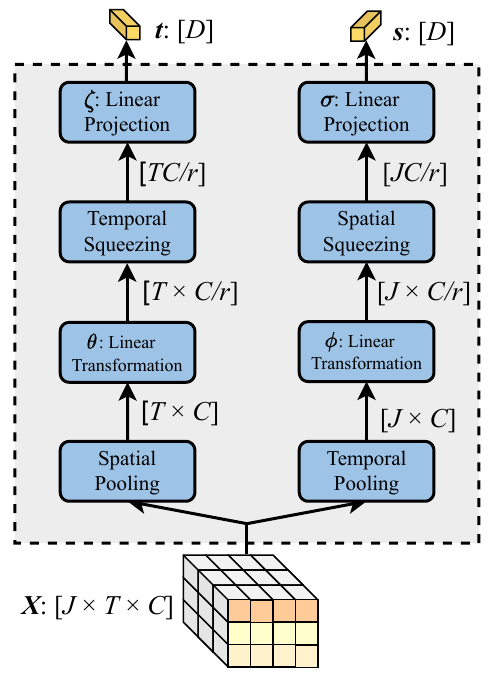}
    \caption{The architecture of the STFD module.}
    \label{fig: STFD}
\end{figure*}

The skeleton sequences of human actions have the property of spatiotemporal coupling, and each dimension in the encoded representation $\boldsymbol{X}$ reflects distinct semantic information: the spatial dimension contains the topological relations, the temporal dimension includes the motion trajectories, and the channel dimension encompasses the richness of features. To explore fine-grained motion patterns along corresponding dimensions, we decompose the representation into the spatial-aware and temporal-aware features.

The details of the proposed STFD are shown in Figure \ref{fig: STFD}. The representation $\boldsymbol{X}$ is fed into two parallel branches for feature decomposition. Taking the spatial feature decomposition (SFD) module as an example, a temporal pooling layer is adopted to obtain the spatial attentive feature $\boldsymbol{X}_s \in \mathbb{R}^{J \times C}$. And the temporal-specific feature $\boldsymbol{X}_t \in \mathbb{R}^{T \times C}$ is obtained through a spatial pooling layer. Then, two linear transformation functions $\theta$ and $\phi$ with a channel reduction rate of $r$ are used to transform the attentive features into neatly compact representations as follows:
\begin{equation}
    \left\{\begin{matrix}
        \boldsymbol{X}_s^{*}=\theta(\boldsymbol{X}_s)=\boldsymbol{X}_s\boldsymbol{W}_{\theta} \in \mathbb{R}^{J \times C/r} \\
        \\
        \boldsymbol{X}_t^{*}=\phi(\boldsymbol{X}_t)=\boldsymbol{X}_t\boldsymbol{W}_{\phi} \in \mathbb{R}^{T \times C/r}
    \end{matrix}\right.
\end{equation}
where $\boldsymbol{W}_{\theta}, \boldsymbol{W}_{\phi} \in R^{C \times C/r}$ are the weights of the spatial and temporal transformation function, respectively. After this, to retain the attentive information, we squeeze the spatial dimension into the channel dimension in the SFD module, and the TFD module is similar. In the end, two liner transformation functions $\zeta$ and $\sigma$ are used to transform the features to the latent space for further contrastive learning, which can be formulated as:
\begin{equation}
    \left\{\begin{matrix}
        \boldsymbol{s}= \zeta(\boldsymbol{s}^{*}) = \boldsymbol{s}^{*}\boldsymbol{W}_{\zeta} \in \mathbb{R}^{D} \\
        \\
        \boldsymbol{t}= \sigma(\boldsymbol{t}^{*}) = \boldsymbol{t}^{*}\boldsymbol{W}_{\sigma} \in \mathbb{R}^{D}
    \end{matrix}\right.
\end{equation}
where, $\boldsymbol{W}_{\zeta} \in \mathbb{R}^{JC/r \times D}$, $\boldsymbol{W}_{\sigma} \in \mathbb{R}^{TC/r \times D}$ are the weights of final projection function. In this way, we decompose the spatial-specific feature $\boldsymbol{s}$ and the temporal-specific feature $\boldsymbol{t}$ from the spatiotemporal representation $\boldsymbol{X}$.

Thus, the spatiotemporal representation extracted from the skeleton encoder is decomposed into two compact and specific features along the respective dimensions.

\subsection{Contrastive Spatiotemporal Representation Enhancement}
\label{CSRE}
\begin{figure}[h]
    \centering
    \includegraphics[scale=0.615]{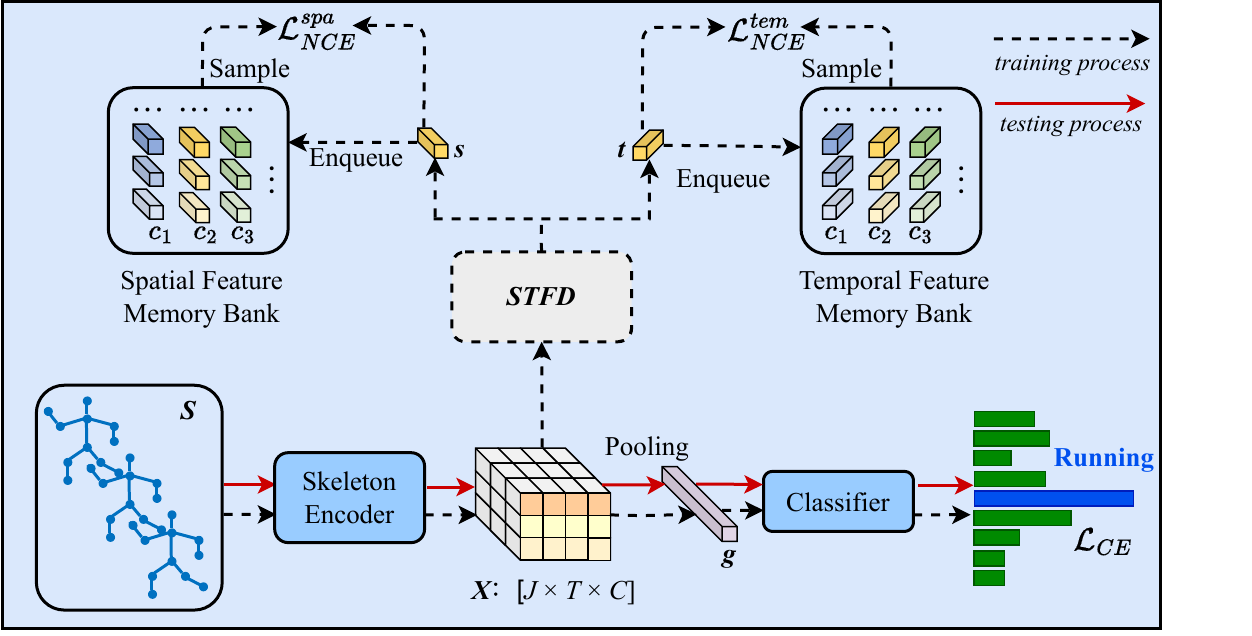}
    \caption{Framework of the proposed CSRE. The black dotted lines represent the training process, and the red solid line represents the testing process. The skeleton encoder extracts spatial-temporal representation $\boldsymbol{X}$ from the input skeleton sequence $\boldsymbol{S}$. Then,} the $\boldsymbol{X}$ is fed into the STFD module to decompose spatial attentive feature $\boldsymbol{s}$ and temporal attentive feature $\boldsymbol{t}$. Two memory banks are built to store the decomposed features respectively. We contrast the positive and negative instance features sampled from the memory banks according to the labels. The memory banks and the contrastive learning process are only used in the training phase and have no impact in the testing phase.
    \label{fig: Framework}
\end{figure}

The framework of the proposed CSRE is shown in Figure \ref{fig: Framework}. To further enhance the distinctiveness of the representation $\boldsymbol{X}$, helping distinguish ambiguous samples, we apply the decomposed components to contrastive learning to explore the cross-sequence semantic relations. The core is to explore the latent data distributions by pulling together the features from the positive pairs and pushing away the negative pairs.

\subsubsection{Memory Bank}
To obtain abundant negative pairs, we construct two memory banks $\mathcal{M}_{spa}$ and $\mathcal{M}_{tem}$ to store the decomposed features. Specifically, to avoid introducing hyperparameters and fully encourage the model to explore the cross-sequence context, we set the two memory banks to store the attentive features of all samples. Thus, the memory \textbf{bank} $\mathcal{M}_{spa}, \mathcal{M}_{tem} \in \mathbb{R}^{L \times D}$. $L$ is the number of the instance of the dataset. When an instance is fed into the network to extract features, the stored features in the memory banks are updated according to the corresponding index.
\begin{equation}
    \left\{\begin{matrix}
        \mathcal{M}_{spa}[i] = \boldsymbol{s}_i \\
        \\
        \mathcal{M}_{tem}[i] = \boldsymbol{t}_i
    \end{matrix}\right.
\end{equation}
where $i$ is the index of the instance in the dataset. Each element in $\mathcal{M}_{spa}$ and $\mathcal{M}_{tem}$ indicates a decomposed feature embedding from an instance. In the memory banks, the corresponding labels with the decomposed features are stored for the feature selection. Particularly, the specific features with the same category label as the current instance are selected as positive pairs, and negative pairs are sampled from the memory banks with other labels.

\subsubsection{Training Objective}
The positive pairs and negative pairs sampled from the two memory banks are used to achieve contrastive learning. To pull together the positive pairs and push away the negative pairs effectively, we define the distances between the two vectors with cosine similarity as follows:
\begin{equation}
    \label{equ: similarity}
    \left \langle \boldsymbol{u}, \boldsymbol{v} \right \rangle = \frac{\boldsymbol{u} \cdot \boldsymbol{v}^{T}}{||\boldsymbol{u}||\;||\boldsymbol{v}||}
\end{equation}
where $\boldsymbol{u}$, $\boldsymbol{v}$ are the features sampled from the memory banks. Then the InfoNCE loss adopted for spatial-specific feature contrasting and temporal-specific feature contrasting can be written as follows:

\begin{equation}
    \mathcal{L}_{NCE}^{spa} = - \sum\limits_{\boldsymbol{s}^{+} \in \mathcal{M}^{+}_{spa}} log \frac{\left \langle \boldsymbol{s}, \boldsymbol{s}^{+}\right \rangle/\tau}{\left \langle\boldsymbol{s}, \boldsymbol{s}^{+} \right \rangle /\tau + \sum\limits_{\boldsymbol{s}^{-} \in \mathcal{M}^{-}_{spa}} \left \langle \boldsymbol{s}, \boldsymbol{s}^{-} \right \rangle /\tau}
\end{equation}
\begin{equation}
    \mathcal{L}_{NCE}^{tem} = - \sum\limits_{\boldsymbol{t}^{+} \in \mathcal{M}^{+}_{tem}} log \frac{\left \langle \boldsymbol{t}, \boldsymbol{t}^{+}\right \rangle/\tau}{\left \langle \boldsymbol{t}, \boldsymbol{t}^{+}\right \rangle/\tau + \sum\limits_{\boldsymbol{t}^{-} \in \mathcal{M}^{-}_{tem}} \left \langle \boldsymbol{t}, \boldsymbol{t}^{-}\right \rangle/\tau}
\end{equation}
where $\tau$ is a temperature hyperparameter in contrastive learning. $\mathcal{M}^{+}_{spa}$ and $\mathcal{M}^{+}_{tem}$ are the sets of the attentive features from the positive samples with the same label as the current feature in the memory banks. $\mathcal{M}^{-}_{spa}$ and $\mathcal{M}^{-}_{tem}$ are the sets of negative features with different lalbes.

Finally, the total training loss function of our proposed CSRE is defined as follows:
\begin{equation}
    \mathcal{L} = \mathcal{L}_{CE} + \mathcal{L}_{NCE}^{spa} + \mathcal{L}_{NCE}^{tem}
\end{equation}

\subsubsection{Sampling Strategy}
To achieve efficient contrast across sequences, the strategy of sampling features from memory banks for contrastive learning has a large impact on the experimental performance. Therefore, in this paper, we follow the hybrid sampling strategy in \cite{huang2023graph}, which combines the hard mining strategy and random sampling strategy. The hard mining strategy is introduced to focus on the hard samples, which means the $N^{+}_H$ hardest positive samples with the lowest similarity and $N^{-}_H$ hardest negative samples with the highest similarity in equation \ref{equ: similarity}. Furthermore, the random sampling strategy is adopted to maintain global random contrast. There are $N^{-}_{R}$ random negative samples selected to contrast with the input instance. In total, the sampling strategy adopted in our CSRE takes contrast efficiency into account and contributes to the hard examples. More details about the sampling strategy are in Section \ref{ablation_study}

\section{Experiment}
\subsection{Dataset Settings}
\textbf{NTU RGB+D.} NTU60 \cite{shahroudy2016ntu} is the most widely used dataset for skeleton-based action recognition. It contains 56,880 samples in 60 action classes. These samples are performed by 40 distinct subjects. It has been configured into two benchmarking structures: cross-subject (X-sub) and cross-view (X-view). In the X-sub, sequences from 20 individuals are allocated for training purposes, with the remaining 20 sets of sequences earmarked for validation. Meanwhile, under the X-view, skeleton sequences are divided based on camera views; two camera views are used for training, while the remaining ones are utilized for evaluation.

\begin{table}[t]
    \scriptsize
    \setlength\tabcolsep{2.5pt}
    \caption{Top-1 action classification accuracy comparison ($\%$) with the state-of-the-art methods on NTU 60 dataset. Particularly, the \textit{2S} incidents the ensemble results of \textit{joint} and \textit{bone} streams. And the \textit{4S} incidents the ensemble results of \textit{joint}, \textit{bone}, \textit{joint motion} and \textit{bone motion} streams. * denotes the ensemble results of the streams reported in their papers. Bold and underlined figures indicate the best and second-best results for each dataset, respectively.}
    \resizebox{\textwidth}{!}
    {
        \begin{tabular}{c|cccc|cccc}
            \toprule
            Dataset                               & \multicolumn{8}{c}{NTU 60}                                                                                                                                                                                                                                                                       \\\hline
            Setting                               & \multicolumn{4}{c|}{X-Sub}                  & \multicolumn{4}{c}{X-View}                                                                                                                                                                                                                         \\\hline
            Method/Streams                        & \textit{J}                                  & \textit{B}                      & \textit{2S}                    & \textit{4S}                     & \textit{J}                      & \textit{B}                    & \textit{2S}                               & \textit{4S}                     \\
            \midrule
            ST-GCN \cite{yan2018spatial}          & 81.5                                        & -                               & -                              & -                               & 88.3                            & -                                                                                                           \\
            Shift-GCN \cite{cheng2020skeleton}    & 87.8                                        & -                               & 89.7                           & 90.7                            & 95.1                            & -                             & 96.0                                      & 96.5                            \\
            TA-CNN \cite{xu2022topology}          & 88.9                                        & 89.2                            & 91.0                           & 91.5                            & 94.5                            & 94.1                          & 95.7                                      & 95.9                            \\
            InfoGCN \cite{chi2022infogcn}         & 89.4                                        & 90.6                            & 91.3                           & 92.3                            & 95.2                            & 95.4                          & 96.2                                      & 96.7                            \\
            HD-GCN \cite{lee2023hierarchically}   & -                                           & -                               & 92.4                           & 93.0                            & -                               & -                             & 96.6                                      & 97.0                            \\
            DD-GCN \cite{li2023dd}                & -                                           & 90.5                            & -                              & 92.6                            & -                               & \textbf{95.7}                 & -                                         & 96.9                            \\
            SelfGCN \cite{wu2024selfgcn}          & -                                           & -                               & -                              & \underline{93.1}                & -                               & -                             & -                                         & 96.6                            \\
            STC-Net\cite{lee2023leveraging}       & -                                           & -                               & \underline{92.5}               & 93.0                            & -                               & -                             & \textbf{96.7}                             & \underline{97.1}                \\
            \hline
            HCN \cite{li2018co}                   & -                                           & -                               & 84.3*                          & -                               & -                               & -                             & 89.9*                                     & -                               \\
            HCN \textit{w/}CSRE                   & -                                           & -                               & $\text{85.1}^{\uparrow{0.8}}$  & -                               & -                               & -                             & $\text{91.2}^{\uparrow{1.3}}$             & -                               \\
            \hline
            2S-AGCN \cite{shi2019two}             & 88.9                                        & 89.2                            & 91.0                           & 91.5                            & 94.5                            & 94.1                          & 95.7                                      & 95.9                            \\
            2S-AGCN \textit{w/}CSRE               & $\text{89.4}^{\uparrow{0.5}}$               & $\text{89.8}^{\uparrow{0.6}}$   & $\text{91.5}^{\uparrow{0.5}}$  & $\text{91.9}^{\uparrow{0.4}}$   & $\text{94.7}^{\uparrow{0.2}}$   & $\text{94.6}^{\uparrow{0.5}}$ & $\text{96.1}^{\uparrow{0.4}}$             & $\text{96.3}^{\uparrow{0.4}}$   \\
            \hline
            CTR-GCN \cite{chen2021channel}        & 89.8                                        & 90.2                            & 92.0                           & 92.4                            & 94.8                            & 94.8                          & 96.3                                      & 96.8                            \\
            CTR-GCN \textit{w/}CSRE               & $\text{90.4}^{\uparrow{0.6}}$               & $\text{90.8}^{\uparrow{0.6}}$   & $\text{92.2}^{\uparrow{0.2}}$  & $\text{92.7}^{\uparrow{0.3}}$   & $\text{95.2}^{\uparrow{0.4}}$   & $\text{95.0}^{\uparrow{0.2}}$ & $\text{\underline{96.6}}^{\uparrow{0.3}}$ & $\text{97.0}^{\uparrow{0.2}}$   \\
            \hline
            Hyperformer \cite{zhou2022hypergraph} & 90.3                                        & 91.1                            & 92.0                           & 92.7                            & 94.5                            & 94.4                          & 95.5                                      & 96.2                            \\
            Hyperformer \textit{w/}CSRE           & $\text{\underline{90.8}}^{{\uparrow{0.5}}}$ & $\textbf{91.3}^{\uparrow{0.2}}$ & $\text{92.3}^{\uparrow{0.3}}$  & $\text{92.9}^{\uparrow{0.2}}$   & $\text{95.0}^{\uparrow{0.5}}$   & $\text{94.7}^{\uparrow{0.3}}$ & $\text{96.0}^{\uparrow{0.5}}$             & $\text{96.6}^{\uparrow{0.4}}$   \\
            \hline
            BlockGCN \cite{zhou2024blockgcn}      & 90.9                                        & -                               & -                              & \underline{93.1}                & \underline{95.4}                & -                             & -                                         & 97.0                            \\
            BlockGCN \textit{w/CSRE}              & $\textbf{91.3}^{\uparrow{0.4}}$             & $\underline{91.2}$              & \textbf{92.6}                  & $\textbf{93.3}^{\uparrow{0.2}}$ & $\textbf{95.6}^{\uparrow{0.2}}$ & \underline{95.4}              & \textbf{96.7}                             & $\textbf{97.2}^{\uparrow{0.2}}$ \\
            \bottomrule
        \end{tabular}
    }
    \label{tab:sota ntu60}
\end{table}

\textbf{NTU RGB+D 120.} NTU120 dataset \cite{liu2019ntu} adds 57367 new skeleton sequences and 60 new action classes to the original NTU60 dataset. There are 32 various configurations in it, each of which depicts a different location and background. The authors offered the cross-subject (X-sub) and cross-setup (X-set) as two benchmark evaluations. In the X-sub, sequences from 53 subjects are for training, and the other 53 subjects are for testing. In the X-set, skeleton sequences are split by setup ID. Samples from even set-up IDs are used for training, and the odds are used for evaluation.

\textbf{Northwestern-UCLA.} NW-UCLA dataset \cite{wang2014cross} contains 1494 video clips of 10 different actions captured from three Kinect cameras. We follow the same evaluation protocol in \cite{wang2014cross}: the first two cameras for training and the other for testing.

\begin{table}[t]
    \scriptsize
    \setlength\tabcolsep{2.5pt}
    \caption{Top-1 action classification accuracy comparison ($\%$) with the state-of-the-art methods on NTU120 dataset. Particularly, the \textit{2S} incidents the ensemble results of \textit{joint} and \textit{bone} streams. And the \textit{4S} incidents the ensemble results of \textit{joint}, \textit{bone}, \textit{joint motion} and \textit{bone motion} streams. * denotes the ensemble results of the streams reported in their papers. Bold and underlined figures indicate the best and second-best results for each dataset, respectively.}
    \resizebox{\textwidth}{!}
    {
        \begin{tabular}{c|cccc|cccc}
            \toprule
            Dataset                               & \multicolumn{8}{c}{NTU 120}                                                                                                                                                                                                                                                                                                               \\\hline
            Setting                               & \multicolumn{4}{c|}{X-Sub}      & \multicolumn{4}{c}{X-Set}                                                                                                                                                                                                                                                                               \\\hline
            Method/Streams                        & \textit{J}                      & \textit{B}                                & \textit{2S}                               & \textit{4S}                               & \textit{J}                                & \textit{B}                                & \textit{2S}                     & \textit{4S}                               \\
            \midrule
            Shift-GCN \cite{cheng2020skeleton}    & 80.9                            & -                                         & 85.3                                      & 85.9                                      & 83.2                                      & -                                         & 86.6                            & 87.6                                      \\
            TA-CNN \cite{xu2022topology}          & 84.0                            & 85.1                                      & 87.8                                      & 88.2                                      & 85.3                                      & 86.3                                      & 89.0                            & 89.6                                      \\
            InfoGCN \cite{chi2022infogcn}         & 84.2                            & 86.9                                      & 88.2                                      & 89.2                                      & 86.3                                      & 88.5                                      & 89.4                            & 90.7                                      \\
            HD-GCN \cite{lee2023hierarchically}   & -                               & -                                         & 89.1                                      & 89.7                                      & -                                         & -                                         & 90.6                            & 91.2                                      \\
            DD-GCN \cite{li2023dd}                & -                               & 86.1                                      & -                                         & 88.9                                      & -                                         & 87.6                                      & -                               & 90.2                                      \\
            SelfGCN \cite{wu2024selfgcn}          & -                               & -                                         & -                                         & 89.4                                      & -                                         & -                                         & -                               & 90.1                                      \\
            STC-Net \cite{lee2023leveraging}      & -                               & -                                         & 89.3                                      & 89.9                                      & -                                         & -                                         & \underline{90.7}                & 91.3                                      \\
            \hline
            2S-AGCN \cite{shi2019two}             & 83.8                            & 84.9                                      & 87.7                                      & 88.1                                      & 85.3                                      & 86.3                                      & 89.0                            & 89.6                                      \\
            2S-AGCN \textit{w/}CSRE               & $\text{84.1}^{\uparrow{0.3}}$   & $\text{85.1}^{\uparrow{0.2}}$             & $\text{87.9}^{\uparrow{0.2}}$             & $\text{88.4}^{\uparrow{0.3}}$             & $\text{86.0}^{\uparrow{0.7}}$             & $\text{86.6}^{\uparrow{0.3}}$             & $\text{89.5}^{\uparrow{0.5}}$   & $\text{90.0}^{\uparrow{0.4}}$             \\
            \hline
            CTR-GCN \cite{chen2021channel}        & 84.9                            & 85.7                                      & 88.7                                      & 88.9                                      & 86.7                                      & 87.5                                      & 90.1                            & 90.5                                      \\
            CTR-GCN \textit{w/}CSRE               & $\text{85.6}^{\uparrow{0.7}}$   & $\text{86.7}^{\uparrow{1.0}}$             & $\text{89.3}^{\uparrow{0.6}}$             & $\text{89.5}^{\uparrow{0.6}}$             & $\text{87.1}^{\uparrow{0.4}}$             & $\text{88.2}^{\uparrow{0.7}}$             & $\text{90.6}^{\uparrow{0.5}}$   & $\text{91.0}^{\uparrow{0.5}}$             \\
            \hline
            Hyperformer \cite{zhou2022hypergraph} & 86.1                            & 87.4                                      & 88.9                                      & 89.9                                      & 87.8                                      & 89.0                                      & 90.6                            & 91.2                                      \\
            Hyperformer \textit{w/}CSRE           & $\text{86.8}^{\uparrow{0.7}}$   & $\underline{\text{88.1}}^{\uparrow{0.7}}$ & $\underline{\text{89.4}}^{\uparrow{0.5}}$ & $\underline{\text{90.3}}^{\uparrow{0.4}}$ & $\underline{\text{88.3}}^{\uparrow{0.5}}$ & $\underline{\text{89.2}}^{\uparrow{0.2}}$ & $\textbf{91.0}^{\uparrow{0.4}}$ & $\underline{\text{91.6}}^{\uparrow{0.4}}$ \\
            \hline
            BlockGCN \cite{zhou2024blockgcn}      & \underline{86.9}                & -                                         & -                                         & \underline{90.3}                          & 88.2                                      & -                                         & -                               & 91.5                                      \\
            BlockGCN \textit{w/CSRE}              & $\textbf{87.2}^{\uparrow{0.3}}$ & \textbf{88.4}                             & \textbf{89.6}                             & $\textbf{90.5}^{\uparrow{0.2}}$           & $\textbf{88.7}^{\uparrow{0.5}}$           & \textbf{89.5}                             & \textbf{91.0}                   & $\textbf{91.8}^{\uparrow{0.3}}$           \\
            \bottomrule
        \end{tabular}
    }
    \label{tab:sota ntu120}
\end{table}

\subsection{Implementation Details}
To fully validate the effectiveness and generalizability of CSRE, we take five different model-based approaches as baseline models. (1) CNN-based method. We select HCN \cite{li2018co} as the CNN-based method to validate CSRE. We reproduce the experimental results using the released code and conduct our CSRE on it. We follow the training recipes described in the paper. (2) GCN-based method. 2S-AGCN \cite{shi2019two} and CTR-GCN \cite{chen2021channel}, the widely used GCN-based model for skeleton-based action recognition. For CTR-GCN, we follow their training strategies. Particularly, for 2S-AGCN, we update the training strategies from CTR-GCN, which can improve the performance significantly. Moreover, to validate whether CSRE is still applicable to the latest models, we also experiment on the combination of proposed CSRE and BlockGCN \cite{zhou2024blockgcn}. (3) Transformer-based method. For transformer-based methods, we choose Hyperformer as the baseline.

\begin{table}[t]
    \footnotesize
    \setlength\tabcolsep{1 pt}
    \centering
    \caption{Top-1 action classification accuracy comparison ($\%$) with the state-of-the-art methods on the NW-UCLA dataset. Particularly, the \textit{2S} incidents the ensemble results of \textit{joint} and \textit{bone} streams. And the \textit{4S} incidents the ensemble results of \textit{joint}, \textit{bone}, \textit{joint motion} and \textit{bone motion} streams. * denotes the ensemble results of the streams reported in their papers. Bold and underlined figures indicate the best and second-best results, respectively. }
    {
        \begin{tabular}{c|cccc}
            \toprule
            Dataset                                    & \multicolumn{4}{c}{NW-UCLA}                                                                                                                                              \\\hline
            \makebox[0.3\textwidth][c]{Method/Streams} & \makebox[0.15\textwidth][c]{\textit{J}} & \makebox[0.15\textwidth][c]{\textit{B}} & \makebox[0.15\textwidth][c]{\textit{2S}} & \makebox[0.15\textwidth][c]{\textit{4S}}  \\
            \bottomrule
            AGC-LSTM \cite{si2019attention}            & -                                       & -                                       & 93.3*                                    & -                                         \\
            DC-GCN+ADG \cite{cheng2020decoupling}      & -                                       & -                                       & 95.3                                     & -                                         \\
            Shift-GCN \cite{cheng2020skeleton}         & 92.5                                    & -                                       & 94.2                                     & 94.6                                      \\
            HD-GCN \cite{lee2023hierarchically}        & -                                       & -                                       & 96.6                                     & 96.9                                      \\
            DD-GCN \cite{li2023dd}                     & -                                       & -                                       & -                                        & 96.7                                      \\
            SelfGCN \cite{wu2024selfgcn}               & -                                       & -                                       & -                                        & 96.8                                      \\
            STC-Net \cite{lee2023leveraging}           & -                                       & -                                       & \underline{96.8}                         & \textbf{97.2}                             \\
            \hline
            2S-AGCN \cite{shi2019two}                  & 92.0                                    & 92.2                                    & 95.0                                     & 95.5                                      \\
            2S-AGCN \textit{w/}CSRE                    & $\text{93.8}^{\uparrow{1.8}}$           & $\text{93.1}^{\uparrow{0.9}}$           & $\text{95.9}^{\uparrow{0.9}}$            & $\underline{\text{97.0}}^{\uparrow{1.5}}$ \\
            \hline
            CTR-GCN \cite{chen2021channel}             & 94.6                                    & 91.8                                    & 94.2                                     & 96.5                                      \\
            CTR-GCN \textit{w/}CSRE                    & $\text{94.8}^{\uparrow{0.2}}$           & $\text{94.6}^{\uparrow{2.8}}$           & $\text{95.7}^{\uparrow{1.5}}$            & $\textbf{97.2}^{\uparrow{0.7}}$           \\
            \hline
            InfoGCN \cite{chi2022infogcn}              & 93.8                                    & 94.2                                    & 95.5                                     & 96.1                                      \\
            Info-GCN \textit{w/}CSRE                   & $\text{94.2}^{\uparrow{0.4}}$           & $\text{95.5}^{\uparrow{1.3}}$           & $\text{95.9}^{\uparrow{0.4}}$            & $\text{96.5}^{\uparrow{0.4}}$             \\
            \hline
            Hyperformer \cite{zhou2022hypergraph}      & 92.7                                    & 95.0                                    & 95.0                                     & 96.6                                      \\
            Hyperformer \textit{w/}CSRE                & $\text{95.3}^{\uparrow{2.6}}$           & $\text{95.3}^{\uparrow{0.3}}$           & $\text{96.3}^{\uparrow{1.3}}$            & $\underline{\text{97.0}}^{\uparrow{0.4}}$ \\
            \hline
            BlockGCN \cite{zhou2024blockgcn}           & \underline{95.5}                        & -                                       & -                                        & 96.9                                      \\
            BlockGCN \textit{w/CSRE }                  & $\textbf{95.9}^{\uparrow{0.4}}$         & \textbf{95.8}                           & \textbf{97.0}                            & $\textbf{97.2}^{\uparrow{0.3}}$           \\
            \bottomrule
        \end{tabular}
    }
    \label{tab:sota nwucla}
\end{table}

For the details of contrastive learning, the size of the established memory banks is the length of the dataset. The dimensions of the attentive features $\boldsymbol{t}$ and $\boldsymbol{s}$ is 256. For all datasets, the number of sampling instances $N^{+}_{H}$, $N^{-}_{H}$ and $N^{-}_{R}$ are set as 128, 512, and 512, respectively. To select the model with the best performance, we experiment with temperatures $\tau$ of 0.5, 0.8, and 1.0. All experiments are conducted with Pytorch \cite{paszke2019pytorch} deep learning library on four GeForce GTX NVIDIA 3090 GPUs. Moreover, to avoid random noise and ensure experiment reproducibility, we adopt the same random seed as the baseline methods and fix it.

\begin{table}[t]
    \footnotesize
    \setlength\tabcolsep{1 pt}
    \centering
    \caption{Top-1 accuracy comparison ($\%$) with recent contrastive learning methods in supervised skeleton-based action recognition on the X-sub benchmark of NTU120 dataset. Bold and underlined figures indicate the best and second-best results, respectively.}
    {
        \begin{tabular}{cc}
            \toprule
            \makebox[0.3\textwidth][c]{Method}                   & \makebox[0.3\textwidth][c]{Acc (\%)} \\
            \midrule
            CTR-GCN \cite{chen2021channel}                       & 84.9                                 \\
            CTR-GCN \textit{w/}GAP \cite{xiang2023generative}    & 85.5                                 \\
            CTR-GCN \textit{w/}SkeletonGCL \cite{huang2023graph} & \underline{85.6}                     \\
            CTR-GCN \textit{w/}FR Head \cite{zhou2023learning}   & 85.5                                 \\
            CTR-GCN \textit{w/}CSRE                              & \underline{85.6}                     \\
            \hline
            Hyperformer \cite{zhou2022hypergraph}                & 86.1                                 \\
            Hyperformer {w/}CSRE                                 & \textbf{86.8}                        \\
            \bottomrule
        \end{tabular}
    }
    \label{tab: contrastive}
\end{table}

\subsection{Comparison With The State-of-The-Art}
We compare our method with previous SOTA methods in Tables \ref{tab:sota ntu60}, \ref{tab:sota ntu120} and \ref{tab:sota nwucla}. To fully validate the proposed CSRE, we report the experimental results in a single modality and the ensemble results of multi-modality. $J$ means the \textit{joint} stream, $B$ denotes the \textit{bone} stream, $2S$ denotes the ensemble results of the above two streams by default, and $4S$ incidents the ensemble results of additional \textit{joint motion} and \textit{bone motion} streams. Particularly, due to the design of their model architecture, the experimental results of VA-LSTM \cite{zhang2017view}, ACG-LSTM \cite{si2019attention} and HCN \cite{li2018co} are the ensemble results of two streams. Therefore, we classify the results into $2S$.

\textbf{Stable Improvements.} As shown in Tables \ref{tab:sota ntu60}, \ref{tab:sota ntu120} and \ref{tab:sota nwucla}, combined with CSRE, all five various baselines achieve solid gains on these five benchmarks of the three datasets over different modalities. Take CTR-GCN on the Xsub benchmark of the NTU120 dataset as a baseline example; it improves by 0.7\% (84.9\% to 85.6\%) in the joint stream, 1.0\% (85.7\% to 86.7\%) in the bone stream.

\textbf{Ensemble Improvements.} As shown in Tables \ref{tab:sota ntu60}, \ref{tab:sota ntu120} and \ref{tab:sota nwucla}, the single-modal performance improvements brought by CSRE can be extended to the results of multi-stream fusion (\textit{2S} and \textit{4S}). It improves CTR-GCN by 0.6\% (88.7\% to 89.3\%) in two streams fusion results and 0.6\% (88.9\% to 89.5\%) in four streams fusion results on the Xsub benchmark of the NTU120.

\textbf{General Applicability.} It can be observed that the proposed CSRE framework is applicable to various baselines, including the recent BlockGCN \cite{zhou2024blockgcn}. The results of CSRE with BlockGCN surpass other experimental results and achieve state-of-the-art techniques.

\subsection{Comparison With Other Contrastive Learning Methods}

We compare our method with the recent contrastive learning methods in supervised skeleton-based action recognition. The comparison is conducted on the X-sub benchmark of the NTU120 dataset. As shown in Table \ref{tab: contrastive}, our CSRE achieves competitive improvements. When conducting experiments based on CTR-GCN \cite{chen2021channel}, the proposed CSRE achieves the same or even higher improvements compared to other contrastive learning methods.

Furthermore, previous contrastive learning methods \cite{xiang2023generative,cheng2020skeleton,zhou2023learning} are only applicable to some GCN-based models due to their design, which limits their generalizability. Specifically, GAP \cite{xiang2023generative} relies on the splitting of spatial dimensions according to human body regions for fine-grained feature alignment, which needs to maintain the spatial dimension in the aggregations. FR-Head \cite{zhou2023learning} is imposed on different stages of GCNs to build a multi-level refinement for stronger supervision, which depends on the design of the mainstream GCN-based method heavily and demonstrates weaker generalization. SkeletonGCL \cite{huang2023graph} suggests that due to the high-order structural information in graphs of the GCN blocks, graph contrast can better benefit graph convolution learning. Thus, SkeletonGCL only incorporates GCN-based methods.

To take advantage of our method, we combine the CSRE with Hyperformer \cite{zhou2022hypergraph}, which is a transformer-based method. Our method outperforms previous contrastive learning methods by 1.2\%, which is a promising result.

\subsection{Ablation Study}
\label{ablation_study}

\begin{figure}[!t]
    \centering
    \includegraphics[scale=0.51]{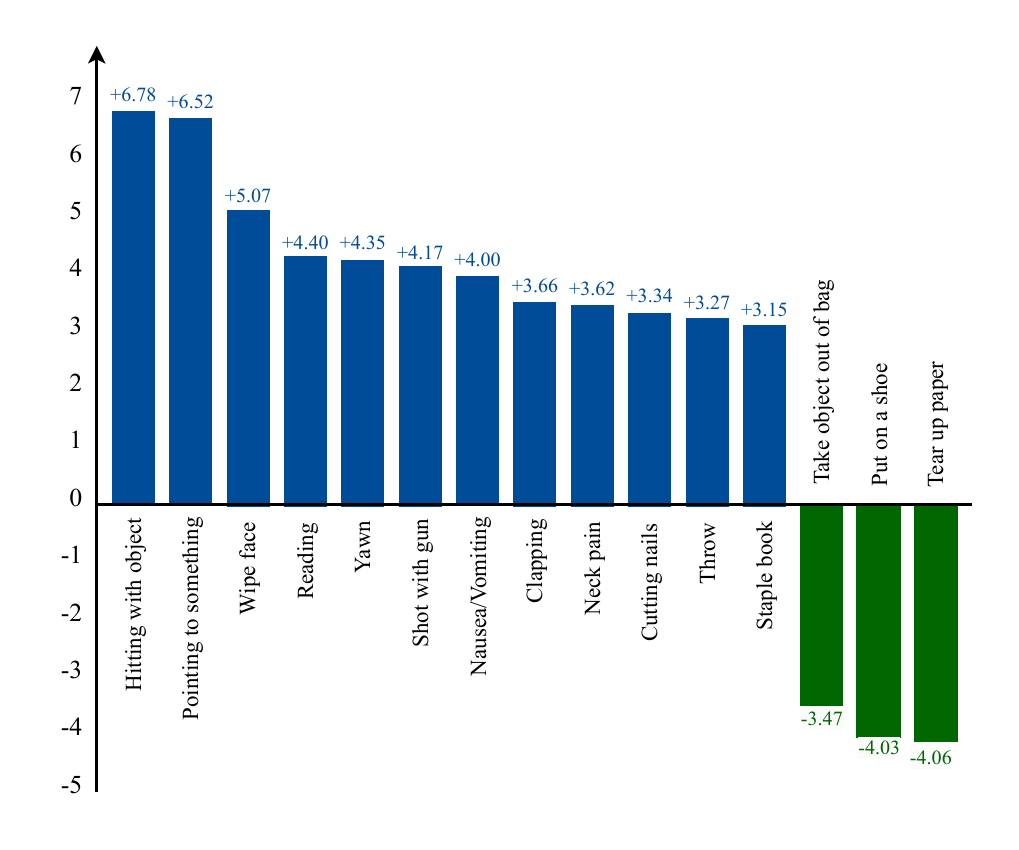}
    \caption{Action class with accuracy differences higher than 3\% between CTR-GCN and our method in the X-sub benchmark of NTU120 dataset.}
    \label{fig: analysis}
\end{figure}

In this section, we evaluate the different experimental settings on the X-sub benchmark of the NTU120 dataset to verify the design of the proposed CSRE.


\textbf{Contrastive Strategy.} In Table \ref{tab: contrastive_strategy}, we conduct experiments to validate the contrastive strategy. We employ the decomposed spatial-aware features or decomposed temporal-aware features to contrastive learning, and these lead to performance improvements, which are owed to the exploration of fine-grained motion patterns along the corresponding dimension. Furthermore, we find that global feature contrast can improve the performance on the baseline with 0.5\% (84.9 \% to 85.6 \%), which owes to the explicit modeling of latent data distributions. Furthermore, we combine the global feature contrast and our CSRE, and the performance drops slightly (85.6\% to 85.5\%), which may be due to that the global features contrast limits the feature decomposition to some extent.

\begin{table}[t]
    \footnotesize
    \setlength\tabcolsep{1 pt}
    \centering
    \caption{Ablation study about the contrastive strategies. SC means applying spatial-specific features to contrastive learning, and TC means applying temporal-specific features to contrastive learning. GC means applying the global feature to contrastive learning. Bold indicates the best results.}
    \begin{tabular}{cc}
        \toprule
        \makebox[0.3\textwidth][c]{Method}      & \makebox[0.3\textwidth][c]{Acc(\%)} \\
        \bottomrule
        CTR-GCN (\textit{w/o} contrast)         & 84.9                                \\
        CTR-GCN \textit{w/}SC                   & $\text{85.4}^{\uparrow{0.5}}$       \\
        CTR-GCN \textit{w/}TC                   & $\text{85.3}^{\uparrow{0.4}}$       \\
        CTR-GCN \textit{w/}GC                   & $\text{85.4}^{\uparrow{0.5}}$       \\
        CTR-GCN \textit{w/}CSRE + \textit{w/}GC & $\text{85.5}^{\uparrow{0.6}}$       \\
        \hline
        CTR-GCN \textit{w/}CSRE                 & $\textbf{85.6}^{\uparrow{0.7}}$     \\
        \bottomrule
    \end{tabular}
    \label{tab: contrastive_strategy}
\end{table}

\textbf{Sampling Strategy.} In Table \ref{tab: sampling_strategy}, we conduct experiments to validate the sampling strategy. In the training phase of the proposed CSRE, we adopt a hybrid sampling strategy to obtain samples from a memory bank. Specifically, hard mining of positive samples and the combination of random sampling and hard mining of negative samples.

\begin{table}[H]
    \footnotesize
    \setlength\tabcolsep{1 pt}
    \centering
    \caption{Ablation study about sampling strategy. \textbf{R}: Random; \textbf{H}: Hard mining. Bold indicates the best results.}
    \begin{tabular}{ccc}
        \toprule
        \multicolumn{2}{c}{Sampling}         & \makebox[0.2\textwidth][c]{\multirow{2}{*}{Acc(\%)}}                 \\
        \makebox[0.2\textwidth][c]{Positive} & \makebox[0.2\textwidth][c]{Negative}                 &               \\
        \midrule
        \multirow{3}{*}{\textbf{R}}          & \textbf{R}                                           & 85.1          \\
                                             & \textbf{H}                                           & 85.2          \\
                                             & \textbf{R} + \textbf{H}                              & 85.4          \\
        \hline
        \multirow{3}{*}{\textbf{H}}          & \textbf{R}                                           & 85.2          \\
                                             & \textbf{H}                                           & 85.3          \\
                                             & \textbf{R} + \textbf{H}                              & \textbf{85.6} \\
        \hline
        \textbf{R} + \textbf{H}              & \textbf{R} + \textbf{H}                              & 85.4          \\
        \bottomrule
    \end{tabular}
    \label{tab: sampling_strategy}
\end{table}

\textbf{Performance on Different Level Actions.} To evaluate the effectiveness of the proposed CSRE, we conduct recognition accuracy for different action categories with different levels of difficulty. Specifically, we gather actions whose accuracy is over 90\% as Easy Level, between 80\% to 90\% as Medium Level, and lower than 80\% as Hard Level for their respective classification results. The experimental results are displayed in Table \ref{tab: acc_levels}. From the results, we can observe that the proposed CSRE achieves relatively significant improvements in the hard and medium-level actions, which include more fine-grained action categories. Moreover, the proposed CSRE has made some gains in the easy-level actions, which is possible thanks to the explicit modeling of the latent data distribution of contrastive learning.

\begin{table}[H]
    \footnotesize
    \setlength\tabcolsep{1 pt}
    \centering
    \caption{Accuracy on different level actions.}
    {
        \ \begin{tabular}{cccc}
            \toprule
            Setting                                    & \multicolumn{3}{c}{Acc(\%)}                                                                                 \\
            \makebox[0.2\textwidth][c]{Method{/}Level} & \makebox[0.15\textwidth][c]{Hard} & \makebox[0.15\textwidth][c]{Medium} & \makebox[0.15\textwidth][c]{Easy} \\
            \midrule
            2S-AGCN                                    & 65.0                              & 86.2                                & 95.1                              \\
            2S-AGCN \textit{w/}CSRE                    & $\text{65.9}^{\uparrow{0.9}}$     & $\text{86.6}^{\uparrow{0.4}}$       & $\text{95.2}^{\uparrow{0.1}}$     \\
            \hline
            CTR-AGCN                                   & 65.9                              & 85.8                                & 95.4                              \\
            CTR-AGCN \textit{w/}CSRE                   & $\text{66.1}^{\uparrow{0.2}}$     & $\text{87.0}^{\uparrow{1.2}}$       & $\text{95.9}^{\uparrow{0.5}}$     \\
            \hline
            Hyperformer                                & 68.3                              & 85.9                                & 95.2                              \\
            Hyperformer \textit{w/}CSRE                & $\text{69.6}^{\uparrow{1.3}}$     & $\text{87.0}^{\uparrow{1.1}}$       & $\text{95.4}^{\uparrow{0.2}}$     \\
            \bottomrule
        \end{tabular}
    }
    \label{tab: acc_levels}
\end{table}

Furthermore, we compare our results with a SOTA model CTR-GCN \cite{chen2021channel}, and the results of class with accuracy differences higher than 3\% between CTR-GCN and our method are displayed in Figure \ref{fig: analysis}. We can observe that actions such as "hitting with object", "pointing to something", "wipe face", "reading" and "yawn" benefit from the proposed CSRE due to the great capacity to obtain discriminative and semantically distinct features. Besides, the reason for the poor performance of actions "take object out of bag", "put on a shoe", and "tear up paper" is that such actions are strongly object-related, making it challenging to recognize and contrast.

\textbf{Training Consumption.} In Table \ref{tab: consumption}, we report the training consumption on NTU120. With our method, the training memory usage only slightly increases with different baseline models, and the increase in training time remains within an acceptable range, which proves the efficiency of the design. Besides, although we set the two memory banks to store the features of all samples, as the size of the dataset increases, it is necessary to balance computational consumption and performance by limiting the capacity of memory banks. Similar to MoCo\cite{he2020momentum}, using a queue-based mechanism to manage the memory bank capacity is a viable approach.
\begin{table}[H]
    \footnotesize
    \setlength\tabcolsep{1 pt}
    \centering
    \caption{The comparison of training consumption.}
    {
        \begin{tabular}{ccc}
            \toprule
            \makebox[0.3\textwidth][c]{Method} & \makebox[0.2\textwidth][c]{Memory-Usage (G)} & \makebox[0.2\textwidth][c]{Time (min/epoch)} \\
            \midrule
            2S-AGCN                            & 7.5                                          & 5.1                                          \\
            2S-AGCN \textit{w/}CSRE            & 7.8                                          & 8.8                                          \\
            \hline
            CTR-GCN                            & 11.5                                         & 19.9                                         \\
            CTR-AGCN \textit{w/}CSRE           & 11.7                                         & 23.5                                         \\
            \hline
            Hyperformer                        & 14.5                                         & 8.9                                          \\
            Hyperformer \textit{w/}CSRE        & 14.9                                         & 12.3                                         \\
            \bottomrule
        \end{tabular}
    }
    \label{tab: consumption}
\end{table}

\subsection{Qualitative Analysis}

\begin{figure}[!t]
    \centering
    \includegraphics[scale=0.45]{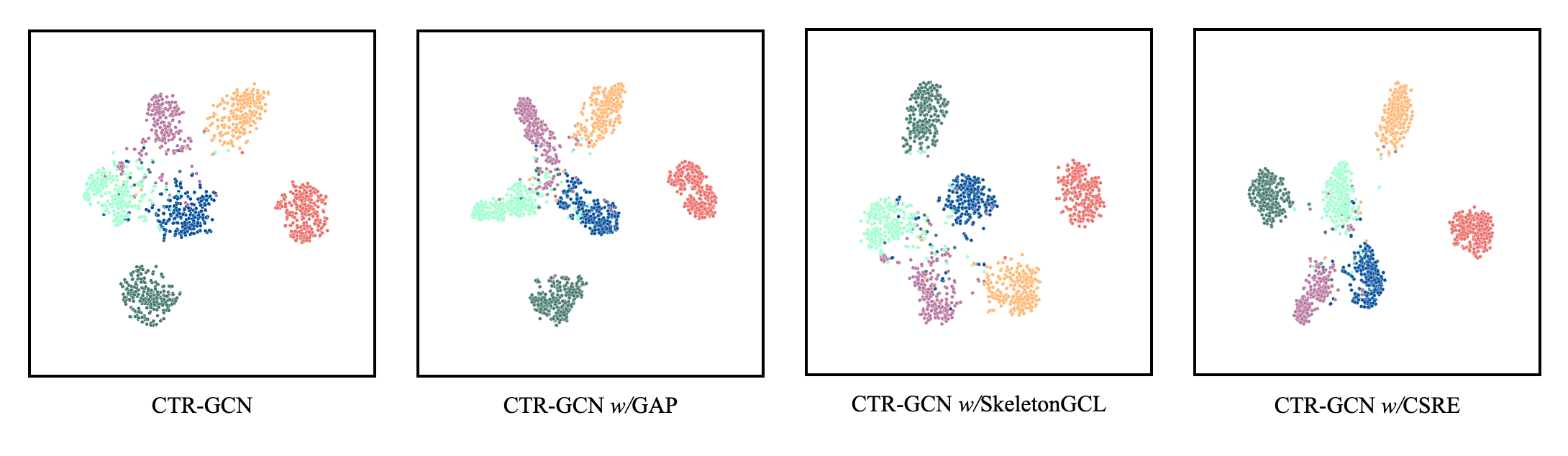}
    \caption{Visualization of feature representation by t-SNE from the sequences in the test set of the X-sub benchmark of NTU120 dataset. Different colors denote different classes.}
    \label{fig: t-SNE}
\end{figure}

In this section, we validate our CSRE through t-SNE \cite{van2008visualizing} distribution visualization of feature representations in the test set of the X-sub benchmark of the NTU120 dataset. As shown in Figure \ref{fig: t-SNE}, all contrastive learning methods, including proposed CSRE, learn more discriminative representations compared to the CTR-GCN \cite{chen2021channel}. Moreover, it can be observed that the feature distribution learned by CSRE has more significant classification boundaries than other contrastive learning methods with fewer ambiguous samples. Thus, the features extracted from the proposed CSRE are more discriminative, which shows that the CSRP improves the feature extraction capacity of the skeleton encoders.

\section{Conclusion}
In this paper, we propose a novel and generically contrastive spatiotemporal representation enhancement (CSRE) framework for skeleton-based action recognition, which can be combined with most previous skeleton encoders. We employ contrastive learning for the spatial-specific and temporal-specific features to encourage the skeleton encoders to enhance the distinctiveness of the representation, which is proven to be effective for recognizing fine-grained action classes and ambiguous samples. The extensive experimental results with five various skeleton encoders on five benchmarks demonstrate the effectiveness of the proposed framework.

\textbf{Limitation and Discussion.} Although the proposed CSRE performs solid improvements on challenging datasets, it also suffers from some dilemmas. As demonstrated in section \ref{ablation_study}, the performer with CSRE is declining in some specific action categories. Since the features cannot form stable clusters during training, CSRE may not have a positive improvement in some categories that originally performed poorly. We suppose this may be due to the poor performance caused by insufficient spatiotemporal modeling of the skeleton encoders. We will do further research on the above aspect in future work.

\section*{Acknowledgments}
This work was supported partly by the National Natural Science Foundation of China (Grant No. 62173045), partly by the Natural Science Foundation of Hainan Province (Grant No. 622RC675), the Fundamental Research Funds for the Central Universities (Grant No. 2020XD-A04-3), and the 2023 “Double First-Class” Discipline Construction Interdisciplinary Team Project of Beijing University of Posts and Telecommunications (Grant No.2023SYLTD02).


\bibliographystyle{elsarticle-num}
\bibliography{egbib}

\clearpage
\clearpage

\end{document}